\documentclass[]{interact}

\usepackage{epstopdf}
\usepackage{hyperref,amsmath,graphicx,amsfonts}
\usepackage{url}
\usepackage{graphicx} 
\usepackage{epsfig}  
\usepackage{amsfonts}
\usepackage{amssymb}
\usepackage{amsmath}
\usepackage{multirow}
\usepackage{booktabs}
\usepackage{url}
\usepackage{caption}
\usepackage{amsfonts}
\usepackage{subcaption}

\numberwithin{equation}{section}

\usepackage{floatrow}
\newfloatcommand{capbtabbox}{table}[][\FBwidth]

\usepackage{blindtext}

\usepackage[longnamesfirst,sort]{natbib}
\bibpunct[, ]{(}{)}{;}{a}{,}{,}

\theoremstyle{plain}

\theoremstyle{definition}

\theoremstyle{remark}

\DeclareMathOperator{\E}{\mathbb{E}}
\begin{document}

	
	\title{Improving Detection of Credit Card Fraudulent Transactions using Generative Adversarial Networks}
	
	\author{
		\name{Hung Ba\textsuperscript{a,b}\thanks{CONTACT Hung Ba Author. Email: hung.nguyen@ed.ac.uk}
		}
		\textsuperscript{a}School of Knowledge Science, Japan Advanced Institute of Science and Technology, Japan\\
		\textsuperscript{b}Business School, The University of Edinburgh, UK\\
	}
	
	\maketitle
	
	\begin{abstract}
		In this study, we employ Generative Adversarial Networks as an oversampling method to generate artificial data to assist on the classification of credit card fraudulent transactions. GANs is a generative model that based on the idea of game theory, in which a generator G and a discriminator D are trying to outsmart each other. The objective of the generator is to confuse the discriminator. The objective of the discriminator is to distinguish the instances coming from the generator and the instances coming from the original dataset. By training GANs on a set of credit card fraudulent transactions, we are able to improve the discriminatory power of classifiers. The experiment results show that the Wasserstein-GAN is more stable in training and produce more realistic fraudulent transactions than the other GANs. On the other hand, the conditional version of GANs in which labels are set by k-means clustering does not necessarily improve the non-conditional versions of GANs. 
	\end{abstract}
	
	\begin{keywords}
		generative adversarial networks; imbalanced learning; creditcard, fraudulent transactions	\end{keywords}

\section{Introduction}
\label{sec:intro}

Credit cards are used as a crucial payment method in modern society, and more fraudulent transactions are increasingly produced in the overwhelming of credit card usages. Fraudulent transactions affect not only the banks and merchants but also the end users because even if they get reimbursement, they could eventually pay more for a higher fee of credit card services.\\

In this study, we employ Generative Adversarial Networks (GANs, \cite{goodfellow_generative_2014}) as an oversampling method to generate artificial data to assist with the classification of credit card fraudulent transactions. GANs is a generative model based on the idea of game theory, in which a generator G and a discriminator D are trying to outsmart each other. The objective of the generator is to confuse the discriminator. The objective of the discriminator is to distinguish the instances coming from the generator and the instances coming from the original dataset. By training GANs on a set of fraudulent transactions and then generating fake fraud transactions to balance the dataset, we compare different oversampling methods on creditcard fraudulent detection.

\section{Literature Review}
\label{sec:lit}
On the distribution of classes, credit granting process and fraud detection are of the sources that produce the highest degree of imbalanced classes. Imbalance dataset (IDS) has almost observations which belong to majority class - good applications and the other belong to a small number of minority class - bad applications. To handle imbalanced dataset, random oversampling the minority class and undersampling the majority class are two common sampling methods. However, oversampling will easily be trapped into overfitting where as undersampling may discard useful data that leads to information loss \cite{he_learning_2009}. As an improvement to random up sampling, Synthetic Minority Over-sampling Technique (SMOTE) \cite{chawla_smote:_2002} synthesizes artificial data in the minority class instead of replication. Random Over-Sampling Examples (ROSE) \cite{menardi_training_2014} generalizes the standard technique of oversampling with replacement the rare examples by allowing the generation of some clones of the observed data, without producing ties.\\

The most recent work of the application of GANs on creating artificial samples to balance the class in classification problems is \cite{douzas_effective_2018}, cGAN performance was evaluated on 71
datasets with different imbalance ratios, number of features and subclustering structures and compared to multiple oversampling methods, using Logistic Regression, Support Vector Machine, Nearest Neighbors, Decision Trees and Gradient Boosting Machine as classifiers. The results show that cGAN performs better compared to the other methods for a variety of classifiers, evaluation metrics and datasets with complex structure. The explanation for this improvement in performance relates to the ability of cGAN to recover the training data distribution, if given enough capacity and training time.

\section{Methodology}
In this section, we provide oversampling background followed by a summary of the GAN, cGAN, WGAN, and WCGAN frameworks following closely the notation in \cite{goodfellow_generative_2014}, \cite{gauthier_conditional_2014}, \cite{arjovsky_wasserstein_2017} and \cite{gulrajani_improved_2017}.
\subsubsection{Oversampling}
Popular oversampling methods including Random Oversampling (ROS), SMOTE, and ADASYN \cite{haixiang_learning_2017} will be used. ROS balances data by randomly duplicating the minority samples. SMOTE selects K nearest neighbors, connects them and forms the synthetic samples. By adaptively changing the weights of the different minority samples to compensate for the skewed distributions, ADASYN uses a density distribution, as a criterion to automatically decide the number of synthetic samples that must be generated for each minority sample. 
\subsubsection{GAN and CGAN}
The generative model $G$, defined as $G:Z\rightarrow X$ where $Z$ is the noise space and $X$ is the data space, aims to capture the real data distribution. The discriminator, defined as $D:X\rightarrow [0,1]$, estimates the probability that a sample came from the  real data distribution rather than the data space generated by $G$. These two models, which could be both multilayer perceptron, compete in a two-player minimax game with value function:
\begin{equation}
\underset{G}{\min\ }\underset{D}{\max}V(D,G)=\underset{x\sim p_{data}}{\E}[logD(x)]+\underset{z\sim p_{z}(z)}{\E}[log(1-D(G(z)))]
\label{eq:1}
\end{equation}

The value $x$ and $z$ are sampled from the real data distribution and noise distribution respectively. The GAN training procedure could be $k$ step(s) for $D$ and one optimizing step for $G$ by means of stochastic gradient descent (SGD).\\ 

Instead of minimizing the probability of generated samples being detected as fake, to prevent vanishing gradient, the optimization process of $G$ could focus on generating as real as possible the noise sample $z$ to confuse $D$ (non-saturating GANs). 
\begin{equation}
	J^G(G) = \underset{z\sim p_{z}(z)}{\E} log(D(G(z)))
\end{equation}


 
The cGAN extends the GAN framework by adding an additional space $Y$ from the real data as follow:\\
$G:Z\times Y \rightarrow X$ and\\
$D:X\times Y \rightarrow [0,1]$\\
And the \ref{eq:1} changes as:
\begin{equation*}
min_Gmax_DV(D,G)=E_D+E_G.
\label{eq:1}
\end{equation*}
where:\\
$E_D=E_{x, y\sim p_{data}(x,y)}[logD(x,y)]$. and\\
$E_G=E_{z\sim p_{z}(z), y\sim p_(y)}[log(1-D(g(z,y), y))]$. \\

The training process of cGAN is almost similar with GAN. By feeding a minibatch of $m$ training examples ${(x_i,y_i)_{i=1}^m}$ and $m$ noise random samples ${z_i}_{i=1}^m$, the logistic cost function for the gradient update of $D$ and $G$ is as follow:
\begin{equation}
J_D=-\dfrac{1}{2m}\Big(\sum_{i=1}^{m}logD(x_i,y_i)+\sum_{i=1}^{m}log\big(1-D(G(z_i,y_i),y_i)\big)\Big).
\end{equation}

\begin{equation}
J_G=-\dfrac{1}{m}\sum_{i=1}^{m}logD(G(z_i,y_i),y_i).
\end{equation}

\subsection{WGAN and WCGAN}
Arovsky et. al., 2017, uses Earth Mover (EM) distance to learn the probability distribution of real data. They propose Wasserstein-GAN (WGAN) to minimize EM distance and the WGAN shows that it could cure the training problem of GANs which requires carefully design of network structures and the balance in training of $D$ and $G$. Specifically, the loss function on training WGAN is:
\begin{equation}
J_D=\dfrac{1}{m}\sum_{i=1}^{m}f_w(x_i)-\sum_{i=1}^{m}f_w(G(z_i)).
\end{equation}
\begin{equation}
J_G=-\dfrac{1}{m}\sum_{i=1}^{m}f_w(G(z_i)).
\end{equation}
where $f$ is a 1-Lipschitz continuous function, parameterized by $w$, that the ``Discriminator'' model need to learn. we could find the details mathematical claims in the original paper of authors.

\subsection{Experimental Setup}
We use credit card transactions data from \cite{pozzolo_calibrating_2015}, which includes a subset of online transactions and consists of 31 coded features. We have 492 frauds out of 284,807 transactions. The dataset is highly imbalanced with the positive class (frauds) accounts for just 0.172\%. The random search is employed for tuning the hyperparameters of GANs frameworks, and the results are reported under 10-fold nested cross-validation (cv).\\ 

The data consists of 31 features: ``time",``amount", ``class" and 28 additional, anonymized features. The class feature is the label indicating whether a transaction is fraudulent or not, with 0 and 1 indicating normal and fraud transaction, respectively. All of the data is numeric and continuous (except the label). The data set has no missing values. \\

For a fast implementation of classification algorithm, we use XGBoost \citep{chen_xgboost:_2016} with max depth equals 4 and area under the curve as an evaluation matric. \\

In this study, we employ GAN as an oversampling method to increase number of the minority class by using the trained, converged generator to create artificial fraud samples. All four GANs models are trained on the full fraud samples, the stopping criteria is defined by manually investigating the loss of both generator and discriminator. We use 10-fold cross validation to examine the quality of generated fraudulent transactions. Let $T_{nk}$ and $T_{fk}$ be the number of normal and fraudulent transactions of fold $k$, respectively. The experiment procedure could be summarized as follows:\\

\noindent \texttt{
	\small
\textbf{for} $k$ in number of folds do:\\
\indent 1.  separate data to training set: $DT_k$ and test set: $T_k$:\\
\indent 2.  \textbf{for} number of training iterations do:
\begin{itemize}
	\item Generate artificial samples $G(z)$.
	\item Train $D$ using both real samples and $G(z)$
	\item Train combined model $G(D(G(z)))$.
	\item Log XGB predictive performance on classifying $G(z)$ and real samples.
	\item Find the iteration $i$ at which XGB performance is lowest.
\end{itemize}
\indent \indent 3. Find the iteration $i$ at which XGB performance is lowest.\\ 
\indent 4. Generate $T_{gfk}^i$ artificial data such that $T_{fk}+T_{gfk}^i = T_{nk}$.\\
\indent 5.   Train and test XGB on the augmented training data $\{DT_k \cup T_{gfk}\}$ and $Test_k$, respectively.\\
}

\subsubsection{Performance Measurements}
To compare seven sampling methods, we train one classifier, Logistic Regression (LR), on balanced data and examine its performance on a separate test set with Area under the ROC curve (AUC), Area under the PR curve (AUPRC), Recall, Precision, and F1-Score. We pay attention to the categorical prediction ability since transaction stagnation causes by misclassifying the normal transactions also threatens the customer relationships of the merchants or financial institutions.

\section{Results}
All four GANs models are trained on 80\% fraud samples, the stopping criterion is defined by investigating the loss of both generator and discriminator, at which we use to generate artificial fraud data.\\

Table \ref{tab:hyp} show the hyperparameters of four GANs frameworks found by random search, it should be noted that we only search for learning rate, drop-out rate, and the number of notes in the 3-layer perceptrons. The network architectures are fixed with three layers and trained with mini-batch of 64 samples, Adam optimizer, and Leaky-Relu activation function ($\alpha=0.2$).

%

\begin{table}[htbp] 
	\small
	\begin{tabular}{lccc}
		& \textbf{Learning} & \textbf{Drop-Out} & \textbf{\#Nodes} \\
		& \textbf{Rate} & \textbf{Rate} &  \\
		\midrule
		GAN & 0.029 & 0.5 & 85 \\
		CGAN & 0.036 & 0.4 & 46 \\
		WGAN & 0.011 & 0.5 & 63 \\
		WCGAN & 0.022 & 0.22 & 5 \\
		\bottomrule
	\end{tabular}
	\caption{Hyperparameters}
	\label{tab:hyp}
\end{table}

\begin{figure}[htbp]
	\centering
	\includegraphics[scale=0.40]{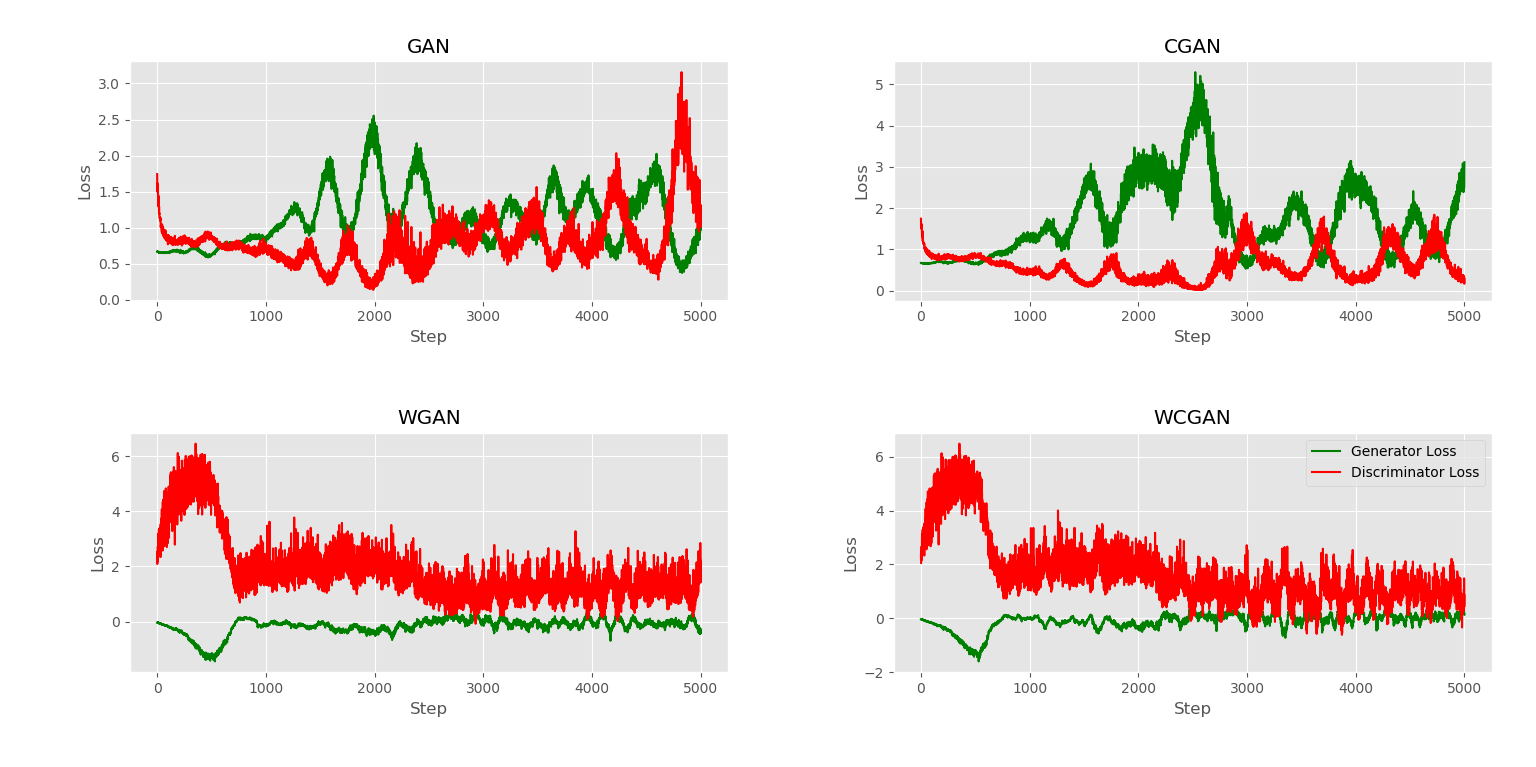}
	\caption{Loss of Generative Adversarial Networks}
	\label{fig:loss}
\end{figure}

Figure \ref{fig:loss} presents the loss of both D and G during the first epoch. While we observe the losses of WGAN variants are stable after 1000 iterations, vanilla GANs (GAN and CGAN) do not actually convert. Hence, we further check for the quality of generated data using Extreme Gradient Boosting Machine (XGB). The step at which we stop training GANs is the step that leads to the lowest accuracy of XGB on discriminating the real and the fake, generated fraud transactions.\\

\begin{figure}[htbp]
	\centering
	\includegraphics[scale=0.68]{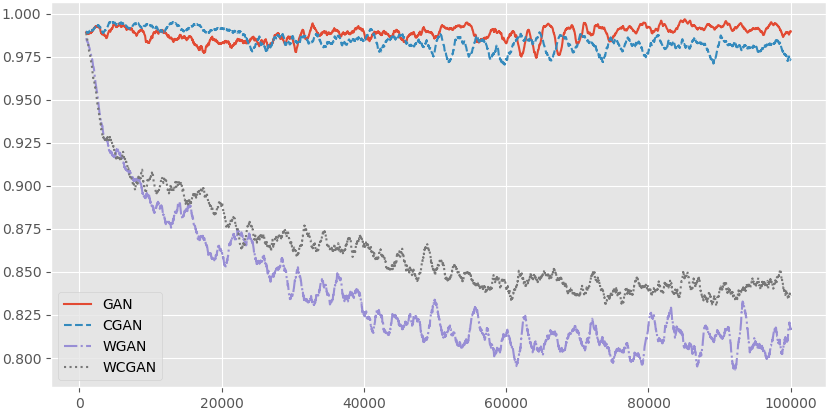}
	\caption{XGB Loss}
	\label{fig:xgbloss}
\end{figure}

From our tests, it appears that our best architecture is the WGAN/WCGAN at the training iteration near 3000 or 5000 as shown in Figure \ref{fig:xgbloss}, 

at the latter step, WGAN/WCGAN achieved an xgboost accuracy of 86\% on detecting fraudulent and generated data (ideally, accuracy would be 50\%). We use all four architectures to generate new fraud data.\\

And to check how efficient the generated data help with detecting fraud credit card transaction, we use up to 80\% of the non-fraudulent data and fraud data. Different amounts of real or generated fraud data are added to this training set, up to 80\% of the fraud data. For the test set, we use the other 20\% of the non-fraud cases and fraud cases. By adding generated data from both an untrained GANs and the best trained GANs to test if the generated data is any better than random noise. 
\begin{figure}[htbp]
	\centering
	\includegraphics[scale=0.48]{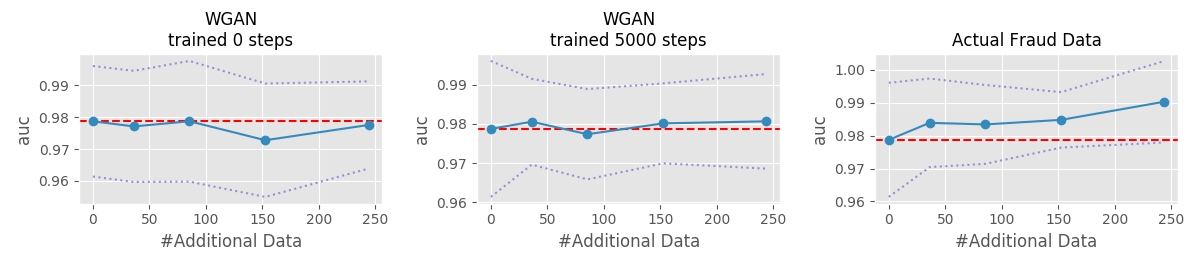}

	\centering
	\includegraphics[scale=0.48]{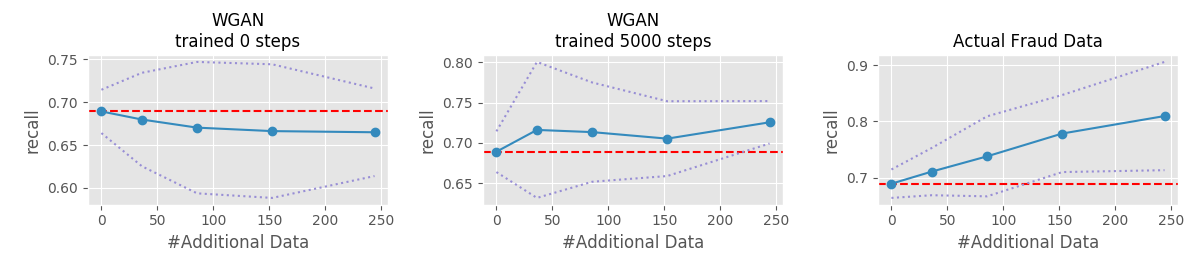}
	
	\centering
	\includegraphics[scale=0.48]{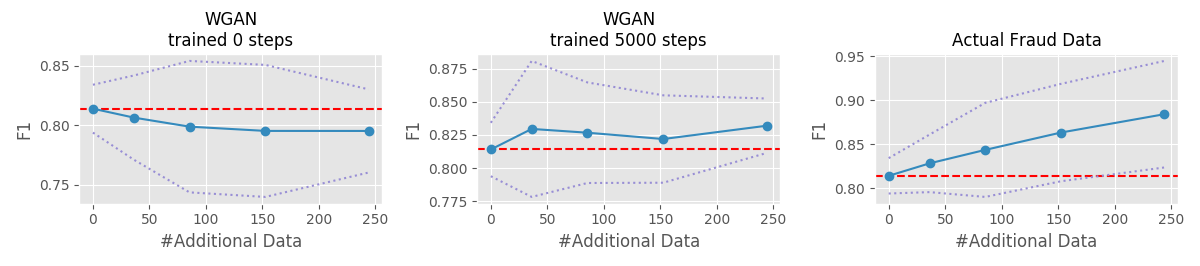}

	\centering
	\includegraphics[scale=0.48]{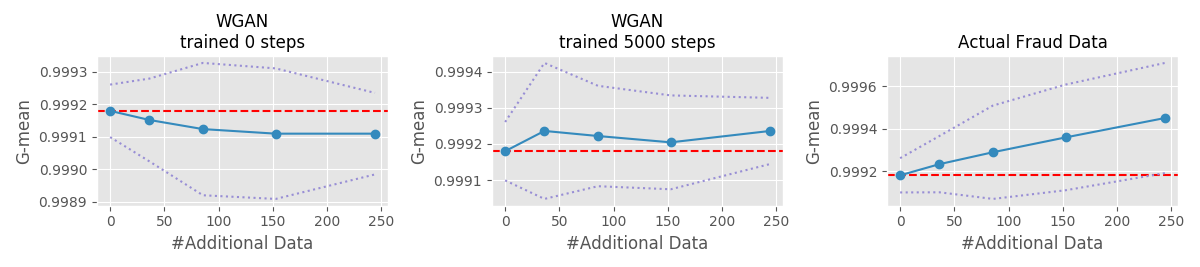}
	\caption{Additional Data vs Outsample Performance}
	\label{fig:addwgan}
\end{figure}

\begin{figure}[htbp]
	\includegraphics[width=0.6\linewidth]{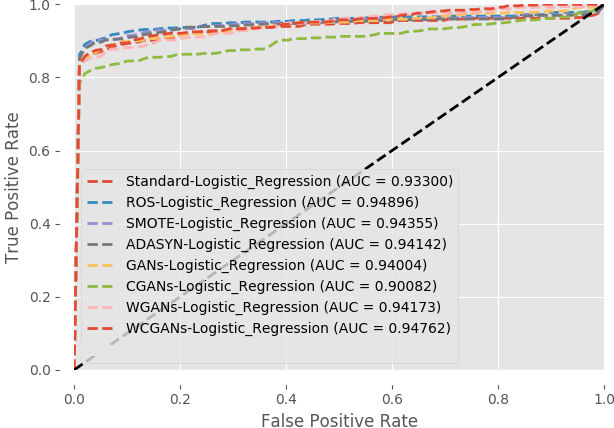}
	\caption{ROC Curves}
	\label{fig:roc}
\end{figure}

Figure \ref{fig:roc} presents all ROC curves of seven balancing methods
and compares them with none sampling setting. As credit card transactions are dominated by the normal transactions, we would focus on the far left-hand side of the ROC curves where ROS and ADASYN are better than the rest. 

The cv performance of LR classifier are shown in Table \ref{tab:per} with five metrics under seven balancing methods. Bold values represent the best. The first row is LR prediction with no sampling and the Rank in the final column is the average rank across five metrics. 
\begin{table}\vspace{1cm}
	\begin{tabular}{lccccc|c}
		&	AUC & AUPRC & Recall & Precison & F1-Score & Rank\\
		\hline
		None  & 0.933 & 0.745 & 0.581 & \textbf{0.908} & 0.680 & 3.8\\
		ROS   & \textbf{0.949} & \textbf{0.750} & 0.882 & 0.067 & 0.123 & 3.2\\
		SMOTE & 0.944 & 0.750 & 0.876 & 0.062 & 0.113 & 4.4\\
		ADASYN & 0.941 & 0.730 & \textbf{0.901} & 0.018 & 0.035 & 5.2\\
		GAN   & 0.940 & 0.637 & 0.502 & 0.777 & 0.501 & 5.6 \\
		CGAN  & 0.901 & 0.631 & 0.564 & 0.643 & 0.444 & 6.4 \\
		WGAN  & 0.942 & 0.723 & 0.803 & 0.500 & 0.583 & 4.2\\
		WCGAN & 0.948 & 0.717 & 0.642 & 0.852 & \textbf{0.710} & 3.2\\
		\hline
	\end{tabular}%
	\caption{Classifier Performance}
	\label{tab:per}
\end{table}\vspace{1cm}

In \emph{AUC}, ROS comes first with WCGAN follows closely. Whereas in \emph{AUPRC}, vanilla balancing methods including the none sampling setting are better than all four GANs frameworks. However, in term of quality and quantity of fraudulent detection, GANs frameworks produce more balancing values in Recall and Precision, which result in better F1-Score.


\section{Conclusions}

\begin{itemize}
	\item Potential application of simple GANs frameworks in enhancing the fraudulent detection in credit card transacions.
	\item GANs are able to learn distributions in situations where the divergence minimization might predict they would fail \cite{fedus_many_2017}.
	\item Wasserstein-GAN is more stable in training and produce more realistic fraudulent transactions than the other GANs.
	\item The conditional version of GANs in which labels are set by k-means clustering does not necessarily improve the non-conditional versions. 
\end{itemize}


\section*{Acknowledgements}

This research is supported by JAIST Off-Campus Research Grant and Doctoral Research Fellow Grant No. 238003. We thank Professor Gary Bolton for his comments when the earlier version of this paper was presented in the Game Theory Conference, Edinburgh, UK, March 2019. All errors retain our own.

\newpage
\bibliography{GAN,Imbalance2,SMEs,Imbalance}

\end{document}